\documentclass[11pt,a4paper]{article}
\usepackage[hyperref]{acl2021}
\usepackage{times}
\usepackage{latexsym}

\usepackage{microtype}

\usepackage{amssymb}
\usepackage{amsmath}
\usepackage{wasysym}
\usepackage{latexsym}

\usepackage{graphicx} % takes care of graphic including machinery

\usepackage{algorithm}
\usepackage[noend]{algpseudocode}

\usepackage{multirow}

\usepackage{silence}
\usepackage{tcolorbox}
\usepackage{xcolor,pifont}
\newcommand*\colourcheck[1]{%
  \expandafter\newcommand\csname #1check\endcsname{\textcolor{#1}{\ding{52}}}%
}
\newcommand*\colourcross[1]{%
  \expandafter\newcommand\csname #1cross\endcsname{\textcolor{#1}{\ding{56}}}%
}
\colourcheck{green}
\colourcross{red}

\usepackage{array}
\newcolumntype{P}[1]{>{\centering\arraybackslash}p{#1}}

\aclfinalcopy % Uncomment this line for the final submission
\usepackage{fancyhdr}
\title{BERT-based distractor generation for Swedish reading comprehension questions using a small-scale dataset}

\author{Dmytro Kalpakchi \\
  Division of Speech, Music and Hearing \\
  KTH Royal Institute of Technology \\
  Stockholm, Sweden \\
  \texttt{dmytroka@kth.se} \\\And
  Johan Boye \\
  Division of Speech, Music and Hearing \\
  KTH Royal Institute of Technology \\
  Stockholm, Sweden \\
  \texttt{jboye@kth.se} \\}

\date{}

\begin{document}
\maketitle

\thispagestyle{fancy}
\begin{abstract}
An important part when constructing multiple-choice questions (MCQs) for reading comprehension assessment are the {\em distractors}, the incorrect but preferably plausible answer options. In this paper, we present a new BERT-based method for automatically generating distractors using only a small-scale dataset. We also release a new such dataset of Swedish MCQs (used for training the model), and propose a methodology for assessing the generated distractors. Evaluation shows that from a student's perspective, our method generated one or more plausible distractors for more than 50\% of the MCQs in our test set. From a teacher's perspective, about 50\% of the generated distractors were deemed appropriate. We also do a thorough analysis of the results.
\end{abstract}

\section{Introduction}
Multiple-choice questions (MCQs) are widely used for student assessments, from high-stakes graduation tests to lower-stakes reading comprehension tests. An MCQ consists of a question (stem), the correct answer (key) and a number of wrong, but plausible options (distractors). The problem of automatically generating stems with a key has received a great deal of attention, e.g., see the survey by \citet{amidei-etal-2018-evaluation}. By comparison, automatically generating distractors is substantially less researched, although \citet{welbl-etal-2017-crowdsourcing} report that manually finding reasonable distractors was the most time-consuming part in writing science MCQs. Indeed, reasonable distractors should be grammatically consistent and similar in length compared to the key and within themselves. 

% TODO: maybe include this link when we say that we're going to generate 3 distractors
%\citet{rodriguez2005three} suggests that the optimal number of distractors is three, and finding a reasonable triple for each stem-key pair seems indeed time-consuming and challenging.

Given the challenges above, we attempt using machine learning (ML) to aid teachers in creating distractors for reading comprehension MCQs. The problem is not new, however most of the prior work has been done for English. In this paper we propose the first such solution for Swedish (although the proposed method is novel even for English, to the best of our knowledge). The key contributions of this work are: proposing a BERT-based method for generating distractors using only a small-scale dataset, releasing SweQUAD-MC\footnote{The dataset and implementation of our models are available in \href{https://github.com/dkalpakchi/SweQUAD-MC}{this GitHub repository}}, a dataset of Swedish MCQs, and proposing a methodology for conducting human evaluation aimed at assessing the plausibility of distractors.

\section{Background}

\begin{table*}[t]
\centering
\begin{tabular}{lccc}
\hline \textbf{Property} & \textbf{Training} & \textbf{Development} & \textbf{Test} \\ \hline
\# of texts & 434 & 64 & 45 \\
\# of MCQs & 962 & 126 & 102 \\
% \# of MCQs per text & $2.2 \pm 1.4$ & $2.0 \pm 1.2$ & $2.3 \pm 1.2$\\
\# of D & $2.1 \pm 0.5$ & $2.1 \pm 0.4$ & $2.0 \pm 0.2$ \\
Len(Text) & $384.9 \pm 330.1$ & $355.1 \pm 233.1$ & $357.9 \pm 254.3$ \\
Len(A) & $4.2 \pm 3.4$ & $4.4 \pm 3.5$ & $4.6 \pm 4.5$ \\
Len(D) & $4.5 \pm 3.9$ & $4.3 \pm 4.0$ & $4.0 \pm 3.7$ \\
$|$Len(A) - Len(D)$|$ & $1.9 \pm 2.4$ & $1.9 \pm 2.3$ & $1.9 \pm 2.9$ \\
\hline
\end{tabular}
\caption{\label{tab:swequad-stats} Descriptive statistics of SweQUAD-MC dataset splits. A denotes the key, D denotes a distractor, Len(X) denotes a length of X in words. $x \pm y$ shows mean $x$ and a standard deviation $y$}
\end{table*}

\subsection{BERT for NLG}
\citet{devlin-etal-2019-bert} introduced BERT as the first application of the Transformer architecture \citep{vaswani2017attention} to language modelling. BERT uses only Transformer's encoder stacks (with multi-head self-attention, MHSA), while the NLG community relies more on Transformer's decoder stacks (with masked MHSA) for text generation, e.g., GPT \citep{radford2018improving}. However, \citet{wang-cho-2019-bert} showed that BERT is a Markov random field, meaning that BERT learns a joint probability distribution over all sentences of a fixed length, and one could use Gibbs sampling to generate a new sentence. The authors compared samples generated autoregressively left-to-right by BERT and GPT, and found the perplexity of BERT samples to be higher than GPT's (BERT samples are of worse quality), but the n-gram overlap between the generated texts and texts from the dataset to be lower (BERT samples are more diverse).

% Highlight the difference in masking to BERT
\citet{liao-etal-2020-probabilistically} show a way to improve BERT's generation capabilities via changing the masking scheme to a probabilistic one at training time. \emph{Probabilistically masked language models} (PMLMs) assume that the masking ratio $r$ for each sentence is drawn from a prior distribution $p(r)$. The authors proposed to train a PMLM with a uniform prior (referred to as u-PMLM). The absence of the left-to-right restriction allows the model to generate sequences in an word arbitrary order. In fact, \citet{liao-etal-2020-probabilistically} propose to generate sentences by randomly selecting the masked position, predicting a token for it, replacing the masked token with the predicted one and repeating the process until no masked tokens are left. The authors showed that the perplexity of the texts generated by u-PMLM is comparable to the ones by GPT.

%A left-to-right autoregressive model can be regarded as a special case of PMLM with $p(r)$ being uniform and only tokens prior to the masked one being available.

\subsection{Convolution partial tree kernels}
\label{subsec:cptk}
As mentioned previously, plausible distractors should be grammatically consistent with the key. Hence, a metric measuring grammatical consistency would be useful both for quantitative evaluation and as a basis for a baseline method. We propose to use convolution partial tree kernels (CPTK) for these purposes. CPTK were proposed by \citet{moschitti2006efficient} for dependency trees and essentially calculate the number of common tree structures (not only full subtrees) between two given trees. However, CPTKs can not handle labeled edges and were applied to dependency trees containing only lexicals. Another solution, proposed by \citet{croce-etal-2011-structured} and used in this article, is to include edge labels, i.e., grammatical relations (GR), as separate nodes. A resulting computational structure is Grammatical Relation Centered Tree (GRCT), which transforms the original dependency tree by making each PoS-tag a child of a GR node and a father of a lexical node. CPTKs can take any non-negative values and are thus hard to interpret. Hence, we use normalized CPTK (NCPTK) shown in Equation \eqref{eq:nctk}, where $K(T_1, T_2)$ is the CPTK applied to the dependency trees $T_1$ and $T_2$.
\begin{equation}\label{eq:nctk}
    \widetilde{K}(T_1, T_2) = \frac{K(T_1, T_2)}{\sqrt{K(T_1, T_1)} \sqrt{K(T_2, T_2)}},
\end{equation}
Evidently, when $T_1$ and $T_2$ are the same, $\widetilde{K}(T_1, T_2)$ equals to 1, which is the highest value it can take.

\section{Data}
We have collected a Swedish dataset, henceforth referred to as \emph{SweQUAD-MC}, consisting of texts and MCQs for the given texts. The dataset was created by three paid linguistics students instructed to pose unambiguous and independent questions. They were also asked to identify the key with at least two distractors, all of which are contiguous phrases in a given text. Additionally, as the distractors were required to be in the same grammatical form as the key (e.g., both in plural), the students were allowed to change the grammatical form of phrases if they constituted plausible distractors after this change. The exact instructions given to the students along with more details on the used texts are provided in Appendix A.

Each datapoint in SweQUAD-MC consists of a base text and an MCQ, i.e. a stem, the key and at least two distractors. The same text can be reused for different MCQs, but the sets of texts in training ($\sim80\%$), development ($\sim10\%$) and test ($\sim10\%$) datasets are disjoint. However, some overlap in sentences is possible, since the texts might come from the same source. Descriptive statistics of all SweQUAD-MC splits is provided in Table \ref{tab:swequad-stats}.

\section{Method}
\begin{figure}[t]
	\centering
	\includegraphics[width=0.32\textwidth]{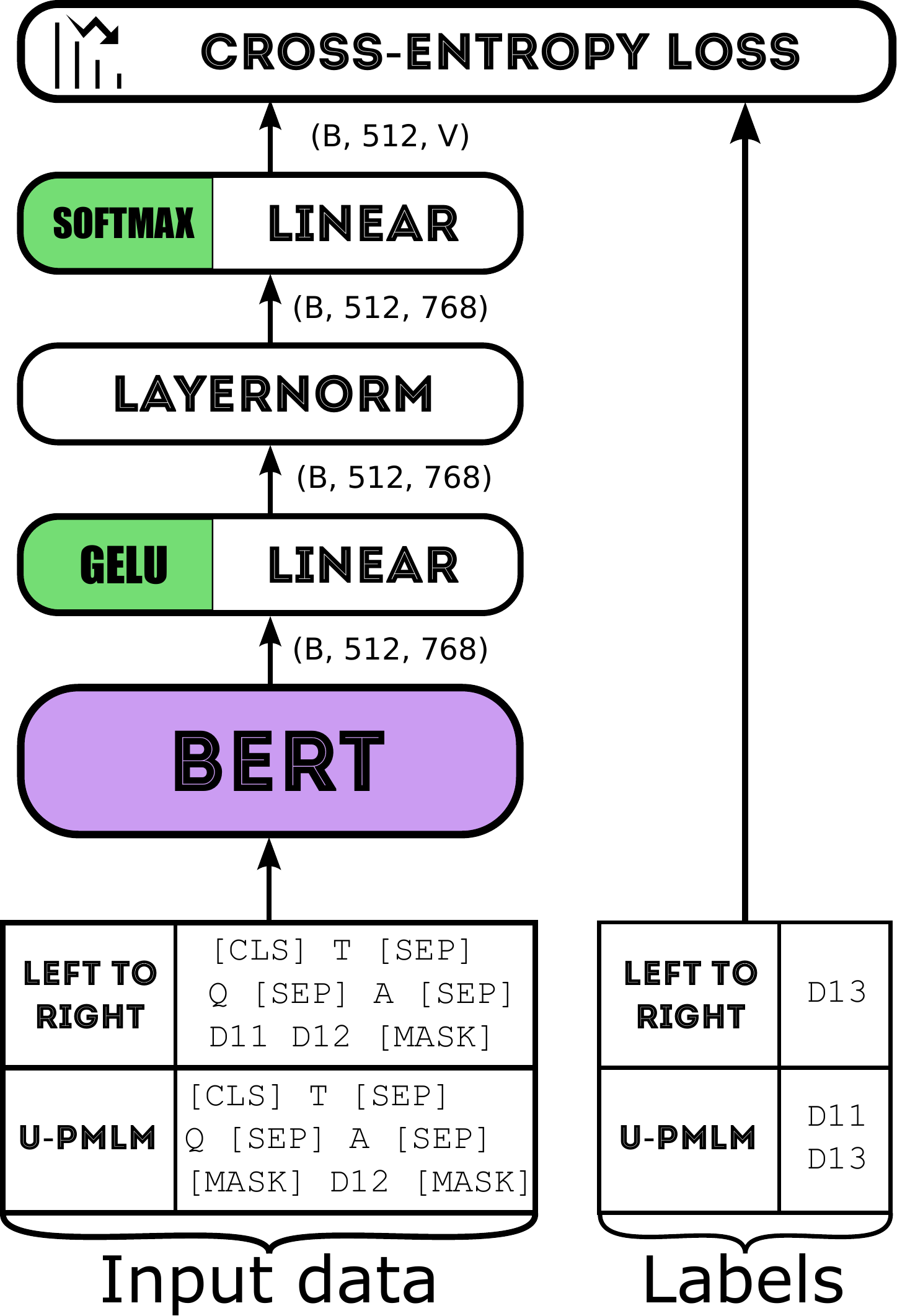}
	\caption{The DG model architecture. B is the batch size and V is the vocabulary size. The light green blocks represent the activation functions for the respective linear layers. The purple block represents parts of the network initialized with the pretrained weights.}
	\label{fig:bdg_architecture}
\end{figure}

Given the small scale of SweQUAD-MC we have decided to fine-tune a pretrained BERT\footnote{bert-base-cased} for Swedish \cite{malmsten2020playing} on the task of distractor generation (DG). For achieving this, we have added on top of BERT two linear layers with layer normalization \cite{ba2016layer} in the middle to be trained from scratch (see architecture in Figure~\ref{fig:bdg_architecture}). The last linear layer is followed by a softmax activation giving probabilities over the tokens in the vocabulary for each position in the text. We trained the model using cross-entropy loss only for tokens in masked positions.

Recall that each MCQ consists of a base text {\tt T}, the stem {\tt Q} based on {\tt T}, the key {\tt A} and (on average) two distractors {\tt D1} and {\tt D2}. The DG problem is then to generate distractors conditioned on the context, consisting of {\tt T}, {\tt Q} and {\tt A}. We provide all context components as input to the BERT model, separated from each other by the special separator token {\tt [SEP]}. Given that BERT's maximum input length is 512 tokens, we trim {\tt T} to the first 384 tokens (later referred to as {\tt T\_384}), since that is the average text length of the training set.

We have explored two different solution variants of DG. The first variant aims at generating distractors autoregressively, left to right. At generation time, the input to BERT consists of a context {\tt CTX} ({\tt T\_384}, {\tt Q} and {\tt A} separated by {\tt [SEP]} token), a {\tt [SEP]} token, and a {\tt [MASK]} token at the end. After a forward pass through BERT, the {\tt [MASK]} token gets replaced by the word with the highest softmax score, which becomes the first word of the first distractor (dubbed {\tt D11}). The generation of the first distractor continues by appending a {\tt [MASK]} token after each forward pass until the network generates a separator token {\tt [SEP]}, which concludes the generation of the first distractor {\tt D1}. The next distractor {\tt D2} is generated in the same way, except that the {\tt CTX} is extended by {\tt D1}. At training time, we use the same procedure, but with teacher forcing, allowing us to use the correct distractor tokens as targets for the cross-entropy loss (see example training datapoints for one MCQ in Table \ref{tab:autoreg-upmlm-ex}).

The second variant is inspired by u-PMLM, and aims at generating distractors autoregressively, but in an arbitrary word order. At generation time, the input to BERT consists of a context {\tt CTX}, a {\tt [SEP]} token, and a predefined number of {\tt [MASK]} tokens (see Section \ref{subsec:quant_eval}). The generation proceeds by unmasking the token at the position where the model is most confident. This differs from unmasking a random position, proposed by \citet{liao-etal-2020-probabilistically}. The training procedure largely follows a masking scheme employed by u-PMLM by drawing the masking ratio from the uniform distribution (see example training datapoints for one MCQ in Table \ref{tab:autoreg-upmlm-ex}). Note that we do not include the {\tt [SEP]} token when training, since we found that the trained model would constantly generate {\tt [SEP]} tokens. 

Each sampled masking ratio $r$ for the u-PMLM variant means that each token in the distractors from the dataset has a probability $r$ to be masked. Hence, different $r$ will potentially result in different number of masked tokens and at different positions. The number of times we draw $r$ per distractor {\tt DX} is proposed to be $\min(\text{Len}({\tt DX}), {\tt MAX\_MASKINGS})$.

% \begin{algorithm}[t]
% \caption{Algorithm for generating in a u-PMLM formulation}
% \label{alg:upmlm}
% \begin{algorithmic}[1]
% \Procedure{GenerateUPMLM}{}
%     \While{any {\tt [MASK]} token exists}
%         \State BertForward()
%         \State pos, token = MaxLogitAcrossPos()
%         \State ReplaceMask(pos, token)
%     \EndWhile
% \EndProcedure
% \end{algorithmic}
% \end{algorithm}

% Freezing weights is bad!
% Describe that we take only first 384 tokens from the text
% Emphasize predicting separators for the autoreg ltr, but not for u-PMLM, since u-PMLM will then predict only [SEP]
% Emphasize that for autoreg generation one generates until [SEP]
% For u-PMLM - one has to decide the number of tokens, which should +/- 1 token compared to CA for distractor generation

\begin{table*}[t]
\centering
\begin{tabular}{l|l}
\hline
\textbf{Input for left-to-right variant} & \textbf{Target}\\ \hline
{\tt [CLS] CTX [SEP] [MASK]} & {\tt D11}\\
{\tt [CLS] CTX [SEP] D11 [MASK]} & {\tt D12} \\
{\tt [CLS] CTX [SEP] D11 D12 [MASK]} & {\tt [SEP]} \\
{\tt [CLS] CTX [SEP] D11 D12 [SEP] [MASK]} & {\tt D21} \\
{\tt [CLS] CTX [SEP] D11 D12 [SEP] D21 [MASK]} & {\tt D22} \\
{\tt [CLS] CTX [SEP] D11 D12 [SEP] D21 D22 [MASK]} & {\tt D23} \\
{\tt [CLS] CTX [SEP] D11 D12 [SEP] D21 D22 D23 [MASK]} & {\tt [SEP]} \\
\hline
\multicolumn{2}{c}{\vspace{1px}}\\
\hline
\textbf{Input for u-PMLM variant} & \textbf{Target(s)}\\ \hline
{\tt [CLS] CTX [SEP] D11 [MASK]} & {\tt D12}\\
{\tt [CLS] CTX [SEP] [MASK] D12} & {\tt D11}\\
{\tt [CLS] CTX [SEP] D11 D12 [SEP] D21 [MASK] [MASK]} & {\tt D22, D23} \\
{\tt [CLS] CTX [SEP] D11 D12 [SEP] D21 [MASK] D23 } & {\tt D22} \\
{\tt [CLS] CTX [SEP] D11 D12 [SEP] [MASK] D22 [MASK]} & {\tt D21, D23} \\
\hline
\end{tabular}
\caption{\label{tab:autoreg-upmlm-ex} Example datapoints extracted from one MCQ if training the autoregressive left-to-right variant (top table) or u-PMLM variant (bottom table). {\tt D1} and {\tt D2} are distractors, assumed to have 2 and 3 words, respectively. {\tt CTX} represents the context, i.e., the sequence {\tt T\_384 [SEP] Q [SEP] A}, where {\tt T\_384} is the first 384 tokens of the text, {\tt Q} is a stem and {\tt A} is the key.}
\end{table*}

\subsection{Baseline}
% TODO: maybe include when describing 3 distractors
% \footnote{Following \citet{rodriguez2005three} who sugessted that 3 is the optimal number of distractors.}
As mentioned in Section \ref{subsec:cptk}, NCPTK measures grammatical consistency between the key and a distractor. Our baseline uses NCPTK on Universal Dependencies (UD) trees \citep{nivre-etal-2020-universal} in the following way. For each given MCQ, we exclude the sentence containing the key from the base text and then parse each remaining sentence $s_i$ of the text, and the key using the UD parser for Swedish. Let $T_{s_i}$ and $T_k$ denote a dependency tree corresponding to $s_i$ and the key respectively. For each $T_{s_i}$, we find all subtrees with the root having the same universal PoS-tag and the same universal features (representing morphological properties of the token) as the root of $T_k$. If no subtrees are found, no distractors can be suggested for this MCQ. Otherwise, we calculate NCPTK between each found subtree and $T_k$ (both as GRCT, but without lexicals). Then we take the textual representation of the $K$ subtrees with the highest NCPTK as the distractor suggestions. 

\section{Experimental setup}
We have used Huggingface's Transformers library \cite{wolf-etal-2020-transformers} for implementing the DG model. The training hardware setup included 16 Intel Xeon CPU E5-2620 v4 (2.10GHz), 64 GB of RAM and 1 NVIDIA GeForce RTX 2080 Ti (11 GB VRAM). For this setup, we have fixed the random seed to 42, the number of training epochs to 6, the batch size to 4 (for both training and dev sets) and {\tt MAX\_MASKINGS} to 20 (for u-PMLM variant only). With these settings, training took about 3.67h for the left-to-right and 3h for the u-PMLM variant. 

UD trees for the baseline were obtained using Stanza package \cite{qi-etal-2020-stanza} and convolution partial tree kernels on the UD trees were calculated using UDon2 library \cite{kalpakchi-boye-2020-udon2}. Baseline requires no training and running our implementation of the baseline takes about a minute on the development or test set.

\section{Evaluation}
Following the analysis of \citet{rodriguez2005three}, we generate three distractors per MCQ for each model. Due to prohibitively high costs of human evaluation, we have divided the evaluation process into two stages. The first stage is quantitative evaluation, which gives limited information about the model's quality, but is sufficient for model selection. The second stage is human evaluation of the best model, selected during the first stage.

\subsection{Quantitative evaluation}
\label{subsec:quant_eval}
Automatic evaluation metrics, such as BLEU \citep{papineni-etal-2002-bleu}, ROUGE \citep{lin-2004-rouge}, METEOR \citep{denkowski-lavie-2014-meteor}, CIDEr \citep{vedantam2015cider}, became popular in NLG in recent years. Essentially, these metrics rely on comparing word overlap between a generated distractor and a reference one. Such metrics can yield a low score even if the generated distractor is valid but just happens to be different from the reference one, or a high score even though the distractor is ungrammatical but happens to have a high word overlap with the reference one (see the article by \citet{callison-burch-etal-2006-evaluating} for a further discussion). Furthermore, they do not take into account how well a generated distractor is aligned with the key grammatically or how challenging the whole group of generated distractors would be. 

To account for the properties mentioned above, we have experimented with a number of quantitative metrics and propose the following set to be used (the whole list is available in Appendix B). In the following list MCQ\% means ``Percentage of MCQ'' and DIS means ``generated distractor(s)''.

\begin{enumerate}
    \item \emph{DisRecall}. Distractor recall.
    \item \emph{AnyDisRefMatch}. MCQ\% with at least 1 DIS matching a reference one.
    \item \emph{AnyDisInText}. MCQ\% with at least 1 DIS appearing in the base text.
    \item \emph{KeyInDis}. MCQ\% with key being among DIS.
    \item \emph{AnySameDis}. MCQ\% with $\geq 2$ identical DIS.
    \item \emph{AllSameDis}. MCQ\% with all identical DIS.
    \item \emph{AnyDisRep}. MCQ\% with $\geq 1$ DIS containing repetitive words contiguously.
    \item \emph{AnyDisEmpty}. MCQ\% with $\geq 1$ DIS being an empty string\footnote{After excluding the special tokens, e.g., {\tt [SEP]}}.
    \item \emph{AnyDisFromTrainDis}. MCQ\% with at least 1 DIS matching with a distractor from training data, but not appearing in the base text.
    \item \emph{MeanNCPTK}, \emph{MedianNCPTK}, \emph{ModeNCPTK}. Mean, median, and mode NCPTK for pairs of UD trees for DIS and keys (all trees as GRCT, but ignoring nodes corresponding to lexicals).
\end{enumerate}

The first group consists of metrics 1-3. The first two metrics count exact matches between generated and reference distractors. The rationale behind metric 3 is our assumption that distractors coming from the same text are more challenging. The higher the values of all these metrics are, the better.

The second group contains metrics 4-8, which give an idea of how challenging the whole group of distractors would be. For instance, duplicate distractors or ones with word repetitions could be excluded by students using common sense. The lower the metrics in this group are, the better.

The third group consists only of metric 9, serving as an overfitting indicator. The metric accounts for the distractors appearing as distractors in training data and high percentage indicates an overfitting possibility. The lower the values, the better. 

% Ideally one would want to include universal features here as well (which one could do instead of lexicals (or along with lexicals). Maybe do now?
The final group (item 10) measures how syntactically aligned generated distractors and the respective keys are. We employ NCPTK to measure the similarity of syntactic structures between each distractor and the respective key. Then we take mean, median and mode of the sequence of NCPTKs obtained in the previous step. The higher the values of these metrics are, the better.

\begin{table}[t]
\centering
\begin{tabular}{l|c|c}
\hline
\textbf{Metric} & \textbf{Baseline} & \textbf{u-PMLM}\\ \hline
DisRecall $\uparrow$ & 1.44\% & 15.31\% \\
AnyDisRefMatch $\uparrow$ & 2.94\% & 26.47\% \\
AnyDisInText $\uparrow$ & 100.0\% & 72.55\% \\
\hline
KeyInDis $\downarrow$ & 0.00\% & 4.9\% \\
AnySameDis $\downarrow$ & 4.9\% & 13.73\% \\
AllSameDis $\downarrow$ & 0.00\% & 1.96\% \\
AnyDisRep $\downarrow$ & 0.00\% & 2.94\% \\
AnyDisEmpty $\downarrow$ & 11.76\% & 0.00\% \\
\hline
AnyDisFromTrainDis $\downarrow$ & NA & 0.98\% \\
\hline
MeanNCPTK $\uparrow$ & 0.43 & 0.43  \\
MedianNCPTK $\uparrow$ & 0.28 & 0.28 \\
\multirow{2}{*}{ModeNCPTK $\uparrow$} & 1.0 & 1.0 \\
& (20.56\%) & (20.69\%) \\
\hline
\end{tabular}
\caption{\label{tab:quant-eval-test} Evaluation of DG models on the test set. When using u-PMLM, shortest distractors were generated first. $\uparrow$ ($\downarrow$) means ``the higher (lower), the better''.}
\end{table}

Based on these metrics, we performed a model selection on the development set and chose the models performing best on the most of these metrics. Left-to-right model generated distractors token by token until either a {\tt [SEP]} token was generated or the length of the distractor was 20 tokens. In contrast, u-PMLM needs the lengths of the distractors to be decided in beforehand, which we set to be the lengths of the two reference distractors and the length of the key\footnote{If reference distractors are not available, we propose to generate distractors with the length differing by at most two words compared to the length of the key.}. Surprisingly, the order of distractors in terms of their length also matters for generation with u-PMLM, so we have tested three options: shortest first, longest first and random order. According to the results of model selection on the development set (presented in detail in Appendix C), u-PMLM models outperformed left-to-right models by a substantial margin. 

The best u-PMLM model (generating shortest distractors first) and the baseline have been evaluated on the test set (see Table \ref{tab:quant-eval-test}). Interestingly, the similarity of syntactic structures between the key and distractors (assessed by NCPTK) is the same for both baseline (that actually relies on NCPTK) and u-PMLM. At the same time, u-PMLM generates more distractors matching the reference ones compared to the baseline (as seen from \emph{DisRecall} and \emph{AnyDisRefMatch}). The baseline generates at least one empty string as a distractor 11.76\% of the time (compared to no such cases for u-PMLM) limiting possibilities of using the baseline in the real-life applications.

\subsection{Human evaluation}
We have used distractors generated on the test set by the best u-PMLM model (selected after quantitative evaluation in Section \ref{subsec:quant_eval}) to conduct human evaluation in 2 stages: from a perspective of a student and a teacher.

\subsubsection{Student's perspective}
\label{subsec:student_eval}
A desirable property of reading comprehension MCQs is that the students should be unable to answer them correctly without reading the actual text. To put more formally, the average number of correctly answered MCQs without reading the actual text (denoted $\overline{N}_s$) should not differ significantly from the average number of correctly answered MCQs when choosing the answer uniformly at random (denoted $\overline{N}_r$). To test for this property, we have formulated the following two hypotheses.\footnote{Preregistration is available \href{https://osf.io/b7pjm}{here}}
\begin{quote}
$\mathcal{H}_0$: $\overline{N}_s = \overline{N}_r$.\\
$\mathcal{H}_1$: $\overline{N}_s \ne \overline{N}_r$.
\end{quote}

For $N$ MCQs with 4 options, $\overline{N}_r = 0.25N$, which for our test set would be equal to $\overline{N}_r = 0.25 \cdot 102 = 25.5$. The appropriate statistical test in this case is one-sample two-tailed t-test with the aim of \emph{not being able to reject} $\mathcal{H}_0$. Given that the purpose is to show that the data supports $\mathcal{H}_0$, we have set both the probability $\alpha$ of type I errors and the probability $\beta$ of type II errors to be $0.05$. Then we have used G*Power \citep{faul2009statistical} to calculate the required sample size for finding a medium effect size (0.5) and the given $\alpha$ and $\beta$, which turned out to be 54 subjects.

Following the calculations above, we have recruited 54 subjects on the Prolific platform\footnote{\href{https://www.prolific.co/}{https://www.prolific.co/}}, and instructed them to choose the most plausible answer to a number of reading comprehension MCQs without providing the original texts. The collected data did not violate any assumptions for a one-sample t-test (see Appendix D.1 for more details). On average, the subjects correctly answered a significantly larger number of questions than $\overline{N}_r$ ($\overline{N}_s=62.26$, $SE=1.09$, $t(53) = 33.51, p < 0.05, r=0.98$). To summarize, the chances of this sample to be collected are very low if $\mathcal{H}_0$ were true.

\begin{figure}
	\centering
	\includegraphics[width=0.4\textwidth]{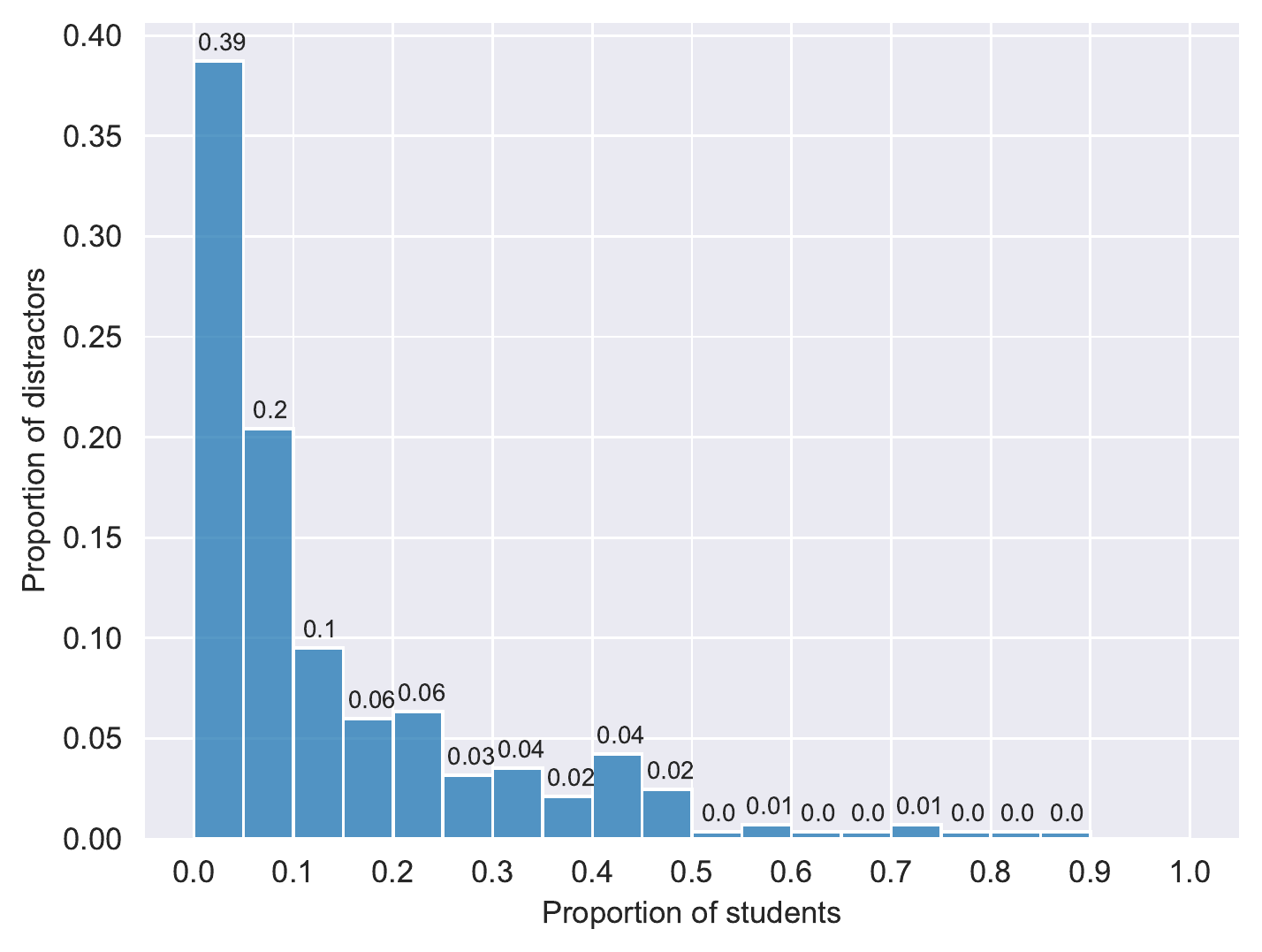}
	\caption{A histogram showing the frequency of choice of distractors in subjects' answers}
	\label{fig:dchoice_freq}
\end{figure}

However, evidently some of the generated distractors were actually plausible, given that $\overline{N}_s \ne N$. To investigate the matter we have plotted the histogram of the frequency of choice of distractors by the subjects in Figure \ref{fig:dchoice_freq}. As suggested by \citet{haladyna1993many}, distractors that are chosen by less than 5\% of students should not be used, which in our case amounts to 39\% of the distractors (the leftmost bar in Figure \ref{fig:dchoice_freq}). If we eliminate these low-frequency distractors (LF-DIS), 68 MCQs (66.67\%) will lose at least one distractor, 10 MCQs (9.8\%) will lose all distractors and thus 34 MCQs (33.33\%) will keep all 3 distractors.

\begin{figure}
	\centering
	\includegraphics[width=0.4\textwidth]{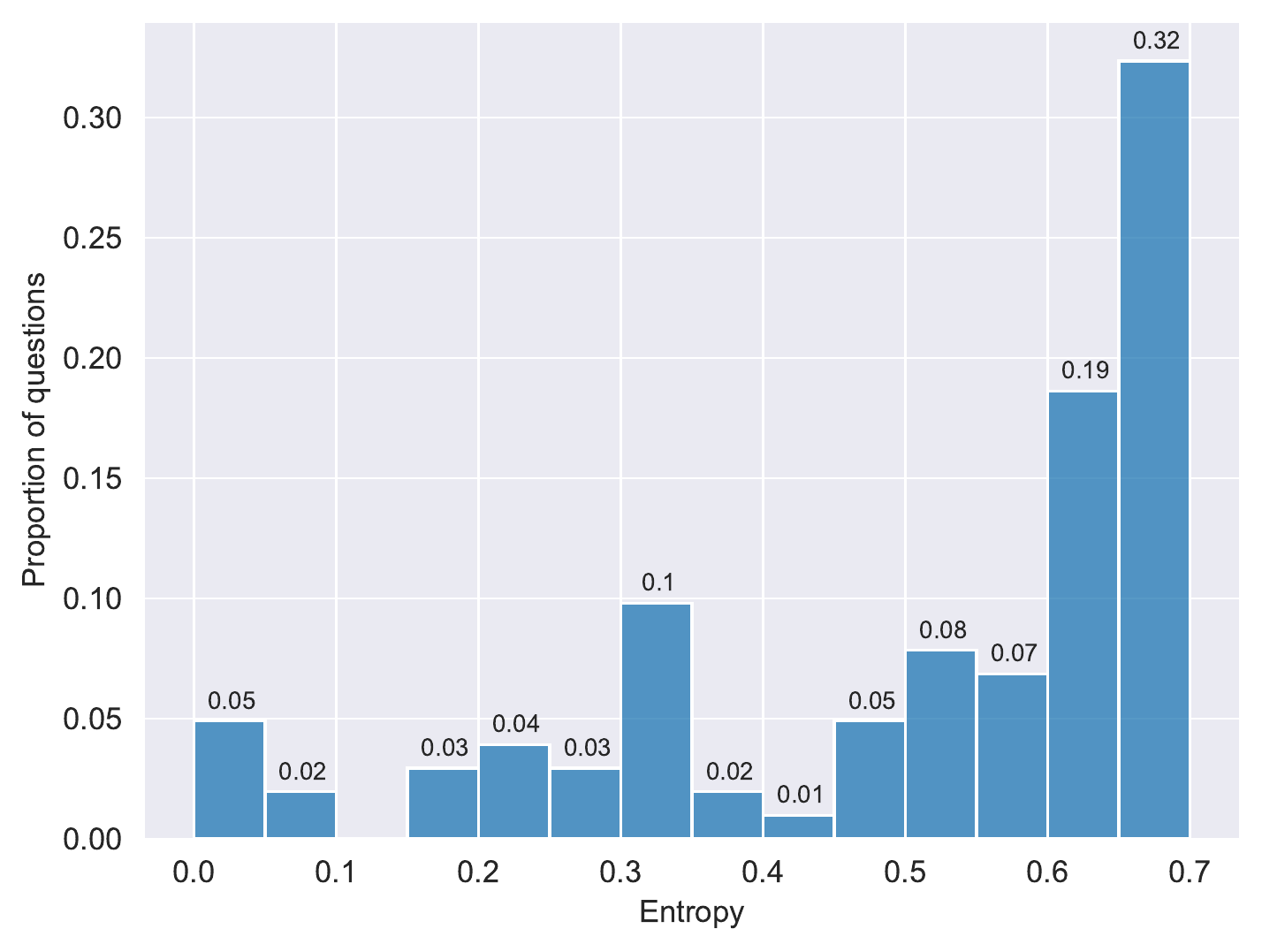}
	\caption{A histogram showing the entropy distribution per question}
	\label{fig:entropy_per_question}
\end{figure}

A more relaxed question is how many MCQs had at least one plausible distractor, which can be estimated by calculating the entropy for each question as shown in Equation \eqref{eq:q_entropy}, where $A$ is the key, $D$ is a distractor, $Q$ is the stem, $P_Q(A)$ ($P_Q(D)$) is the probability that the key (any distractor) is chosen for $Q$ by a subject.
\begin{equation}\label{eq:q_entropy}
    H(Q) = -\sum_{O \in \{A,D\}}p_Q(O)\log(p_Q(O))
\end{equation}
The distribution of entropies per question is shown in Figure \ref{fig:entropy_per_question}. Assuming the natural logarithm, the highest theoretically possible value for $H(Q)$ is $0.69$, if $p_Q(A) = p_Q(D) = 0.5$. 32\% of MCQs had an entropy larger than 0.65, whereas 51\% had an entropy larger than 0.6, which means that half of MCQs had at least one plausible distractor. 

% Explain that entropy of 0 is either all right or all wrong
\subsubsection{Teacher's perspective}
Bearing in mind the findings of Section \ref{subsec:student_eval}, it is interesting to see which of the proposed distractors (especially, among LF-DIS) teachers would mark as acceptable. Given the complexity of such evaluation, using the whole test set was infeasible. To get a representative sample, we used entropy per question (shown in Figure \ref{fig:entropy_per_question}). All MCQs were divided into 5 equally sized buckets by entropy and 9 MCQs were sampled uniformly at random from each bucket, resulting in 45 MCQs in total.

We asked 5 teachers to evaluate each MCQ (presented in a random order for each of them). Each MCQ contained the base text, the stem, the key and the generated distractors. The teachers were instructed to select those of generated distractors (if any) deemed suitable for testing reading comprehension. Additionally, we asked to provide their reasons for each rejected distractor in a free-text input. The inter-annotator agreement (IAA) was estimated using Goodman-Kruskal's~$\gamma$ \cite{goodman1979measures}, specifically its multirater version $\gamma_N$ proposed by \citet{kalpakchi2021quinductor}. On the scale proposed by \citet{rosenthal1996qualitative}, we have found a very large agreement ($\gamma_N=0.85$, see Appendix D.2.2 for more details on IAA calculations).

% To evaluate the inter-annotator agreement (IAA), we have reformulated the problem into a ranking problem, where all accepted distractors were given the rank of 1 and those rejected - the rank of 2. IAA was then estimated using Goodman-Kruskal's~$\gamma$ \cite{goodman1979measures}, specifically its multirater version $\gamma_N$ as proposed by \citet{kalpakchi2021quinductor}. The total number of concordant and discordant pairs were summed for each pair of teachers for each question. The resulting $\gamma_N$ equals to 0.85, indicating a very large agreement on the scale proposed by \citet{rosenthal1996qualitative}.

\begin{figure}[!t]
	\centering
	\includegraphics[width=0.4\textwidth]{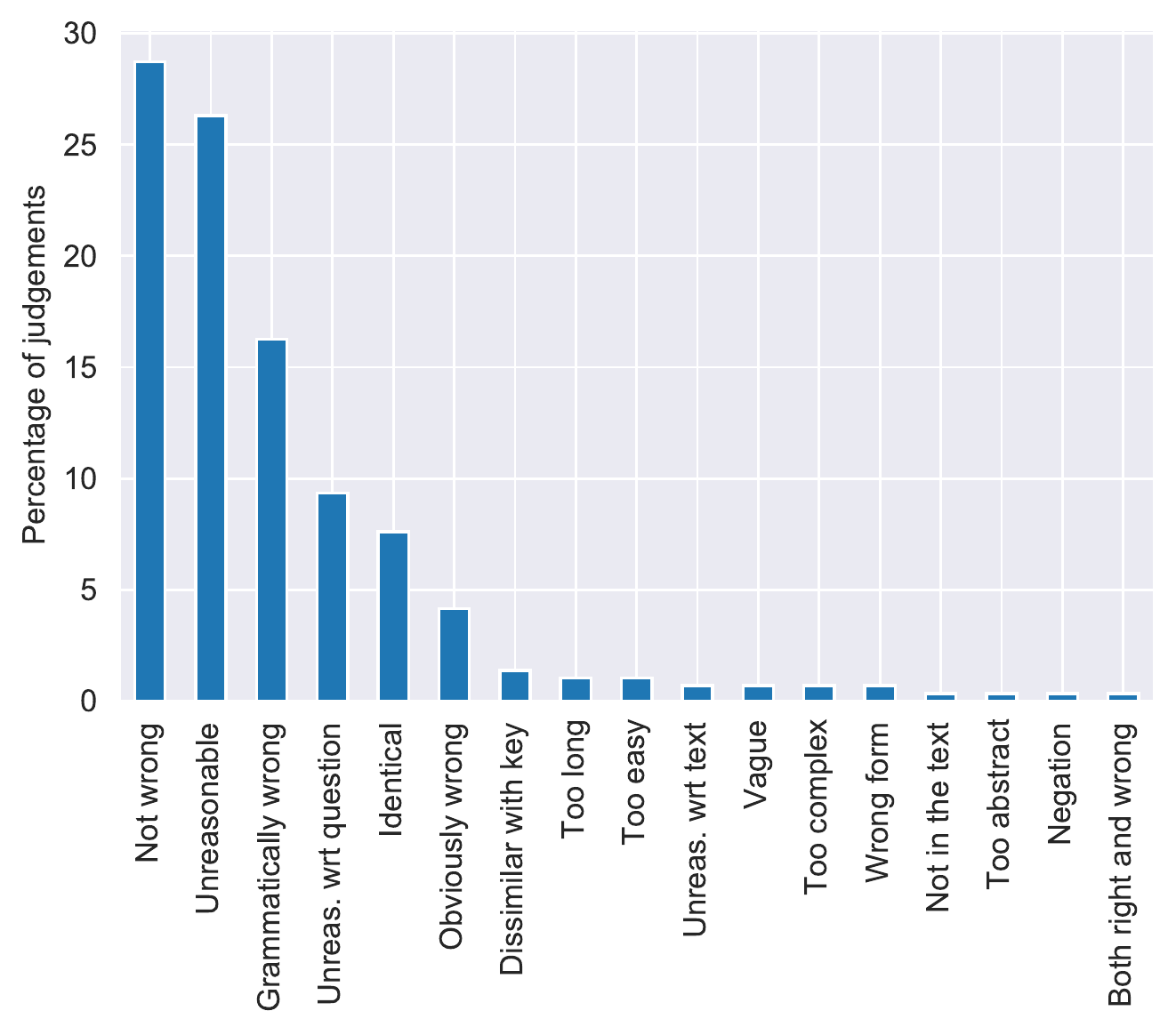}
	\caption{A histogram showing the distribution of teachers' reasons behind rejecting distractors.}
	\label{fig:teachers_rejection_reasons}
\end{figure}

On average, 1.47 distractors per MCQ were accepted by a teacher. Their reasons for rejections are distributed as shown in Figure \ref{fig:teachers_rejection_reasons}. All teachers accepted at least one generated distractor for 39 MCQs (86.7\%), whereas the majority of teachers did so for 27 MCQs (60\%). Interestingly, there are no MCQs in which all 5 teachers have either accepted or rejected all generated distractors. However, the majority of teachers has accepted or rejected all distractors for 4 MCQs (8.9\%) and 6 MCQs (13.3\%) respectively. 

\begin{figure}[!t]
	\centering
	\includegraphics[width=0.4\textwidth]{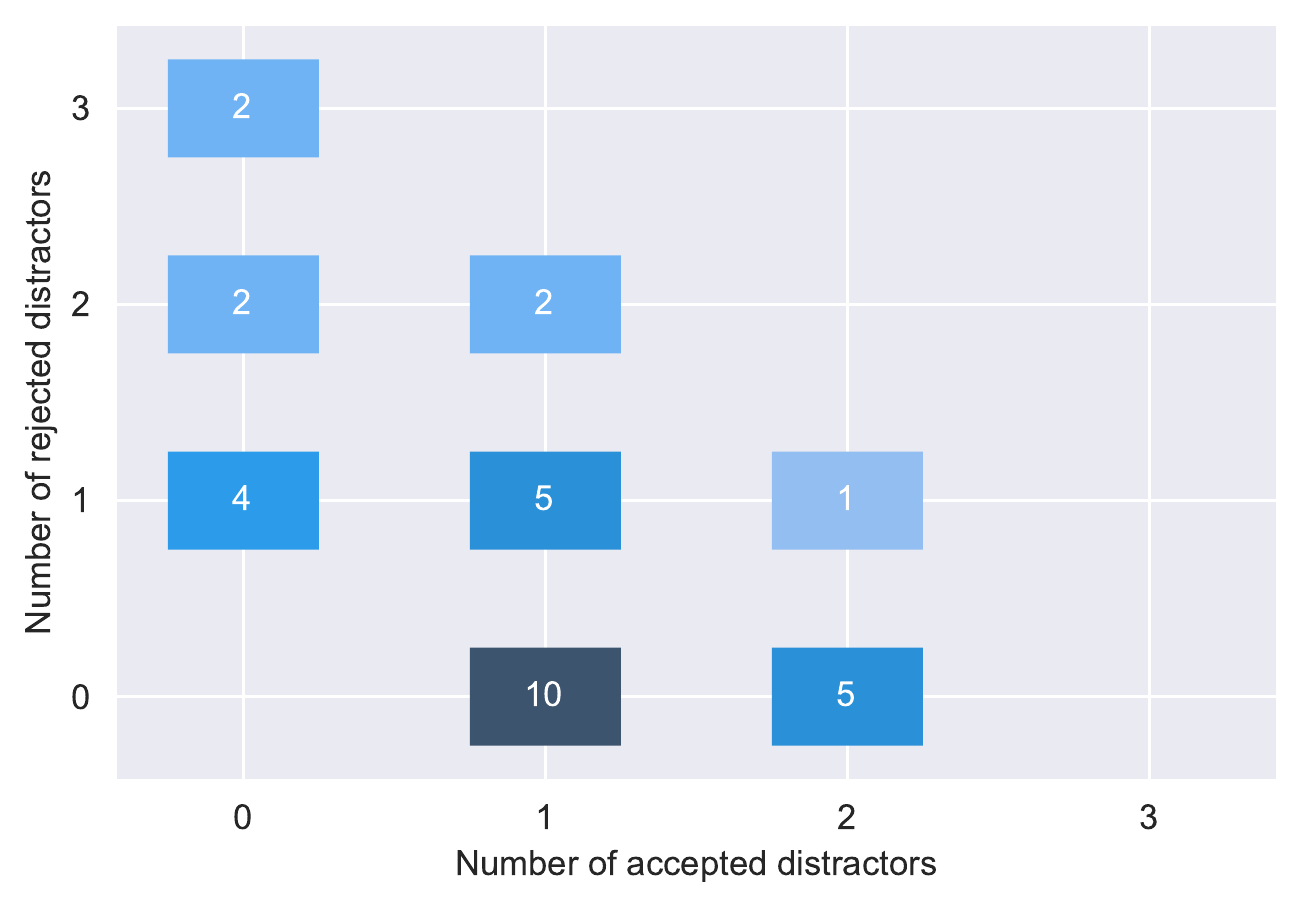}
	\caption{A bi-variate histogram showing the distribution of the 31 MCQs (the numbers on the bars sum to 31) with at least 1 LF-DIS, with respect to their LF-DIS being accepted/rejected by the majority of teachers.}
	\label{fig:teachers_judgement_elminated}
\end{figure}

% GR: "a half" or "half"?
Out of 45 MCQs, 31 (68.9\%) had at least one LF-DIS, as defined in Section \ref{subsec:student_eval}. For these 31 MCQs we report a distribution of accepted/rejected LF-DIS by the majority of teachers in Figure \ref{fig:teachers_judgement_elminated}. Let us call the 15 MCQs with all LF-DIS accepted by the majority of teachers as \emph{mismatch MCQs} (lowest row in Figure \ref{fig:teachers_judgement_elminated}). Interestingly, 12 of the 15 mismatch MCQs had at least one more distractor in addition to LF-DIS being accepted by the majority of teachers. Furthermore, all mismatch MCQs had entropy higher than 0.3. This entails that almost a half of LF-DIS should \emph{not} necessarily be thrown away, since they were accepted by teachers, but the MCQs either happened to have more plausible distractors or subjects might have had relevant background knowledge to answer the questions.

% Let us call the 15 MCQs with all LF-DIS accepted by the majority teachers as \emph{mismatch MCQs} (lowest row in Figure \ref{fig:teachers_judgement_elminated}). The 8 MCQs with all LF-DIS rejected are referred to as \emph{match MCQs} (leftmost column in Figure \ref{fig:teachers_judgement_elminated}). The entropy distribution per question for both groups is presented in Figure \ref{fig:match_mismatch_entropy}. 

% \begin{figure}[!htb]
% 	\centering
% 	\includegraphics[width=0.45\textwidth]{images/match_mismatch_entropy.pdf}
% 	\caption{A histogram showing the entropy distribution per question for match and mismatch MCQs.}
% 	\label{fig:match_mismatch_entropy}
% \end{figure}

% Interestingly, most of the mismatch MCQs have entropy higher than 0.5 (with maximum being 0.69), meaning that the remaining distractors might have been more plausible.

\section{Related work}
We employed a systematic process to get a comprehensive overview of DG methods (see Appendix E for more details). Out of the resulting 28 articles (see an overview in Table \ref{tab:related-survey}), only 2 worked with a language other than English (Chinese and Basque). In this paper we work on reading comprehension MCQs, which makes only 12 papers, dealing with factual questions, relevant.
%To get a comprehensive overview of methods for generating distractors for MCQs, we employed a two-step process. The first step was to issue queries ``distractor generation'' and ``multiple choice question generation'' to ACL Anthology and Google Scholar. The second step was to select relevant references from the ``Related work'' sections of these articles. After filtering out irrelevant results (see more details in Appendix), 28 articles remained (see Table \ref{tab:related-survey} for an overview). Only 2 papers worked with a language other than English (Chinese and Basque). In this paper we work on reading comprehension MCQs, which makes only 12 papers, dealing with factual questions, relevant.

% The result was 20 articles from ACL Anthology and 4 additional ones from Google Scholar.  This resulted into 15 additional articles. Out of found 39 articles, 11 were filtered out (8 focused only on generating questions, 1 relied mostly on expert knowledge, 1 on the auxiliary relation extraction task and 1 was a demo paper), leaving 28 articles in total.

Two of these used rule-based approaches. \citet{majumder-saha-2015-system} generated MCQs for cricket domain and used a number of hand-crafted rules based on gazeteers and Wikipedia entries to generate distractors. \citet{mitkov-ha-2003-computer} proposed to generate distractors for MCQs on electronic instructional documents using WordNet. 

\begin{table}[b]
\centering
\begin{tabular}{lc}
\hline \textbf{Problem/method property} & \textbf{\#} \\ \hline
$\blacksquare$ Extractive & 14\\
$\blacksquare$ Generative, rule-based & 7\\
$\blacksquare$ Generative, neural & 7\\
\hline
$\newmoon$ Only automatic evaluation & 5 \\
$\newmoon$ Only human evaluation & 19 \\
$\newmoon$ Automatic and human evaluattion & 4 \\
\hline
$\blacktriangle$ Cloze-style, single-word answers & 14 \\
$\blacktriangle$ Cloze-style, continue the sentence & 2 \\
$\blacktriangle$ Factual questions & 12 \\
\hline
\end{tabular}
\caption{\label{tab:related-survey} 28 related works broken down by method~($\blacksquare$), type of evaluation ($\newmoon$) and types of questions for which distractors have been generated ($\blacktriangle$)}
\end{table}

Six of these relied on extractive approaches. \citet{liang-etal-2018-distractor}, \citet{welbl-etal-2017-crowdsourcing}, and \citet{ha-yaneva-2018-automatic} formulated choosing a distractor as a ranking problem from the given candidate set. In the first two articles the candidate set constituted all distractors from the available MCQ dataset. The authors then trained ML-based ranker(s) for choosing the best distractors. In the last one, the candidate set was created using content engineers. Distractors with a high similarity of their concept embeddings (summed for multiple words) and appearing in the same document as the key are ranked higher. \citet{stasaski-hearst-2017-multiple} and \citet{araki-etal-2016-generating} worked in the domain of biology. The former used an ontology and the latter employed event graphs containing information about coreferences to generate distractors. \citet{karamanis-etal-2006-generating} used thesaurus and tf-idf to identify key concepts in the given text and then select as distractors those having the same semantic type as the key.

The remaining four employed neural methods and are most relevant among the surveyed. \citet{qiu-etal-2020-automatic} trained a sequence-to-sequence (seq2seq) model with a number of attention layers. \citet{zhou2020co} also employed a seq2seq model, but with a hierarchical attention to capture the interaction between a text and a question, as well as semantic similarity loss. Both articles used a beam search combined with filtering based on Jaccard coefficient at generation time. \citet{offerijns2020better} trained a GPT-2 model to generate 3 distractors for a given MCQ, and used BERT-based question answering model for quantitative evaluation (along with human evaluation). 

Finally, \citet{chung-etal-2020-bert} proposed a BERT-based method for English with answer-negative regularization, penalizing distractors for containing the same words as the key, and training a sequential and a parallel MLM model simultaneously. At generation time, they generate one distractor, and then create a distractor set of the predefined size based on sampling from the probability distribution returned by BERT for each token of the distractor. Then they rank every triple of distractors based on the entropy of a separately trained QA model.

% TODO: add the exact size comparison to RACE
Our method also relies on BERT, but has a number of differences beyond being applied to Swedish. Firstly, we did not include answer-negative regularization, since it is not always a good strategy. For instance, given the stem ``When should you pay a fee if you apply for a visa?'' and a key ``before you have submitted the application'', the best distractor would be ``after you have submitted the application'', which shares most of the words with the key. Secondly, we generate distractors in arbitrary word order compared to left-to-right generation in \cite{chung-etal-2020-bert}. Thirdly, at generation time, we use previously generated distractors as input for generating next ones, and always take tokens with a maximum probability. This lowers the risk of generating ungrammatical distractors. Finally, our training set is 100 times smaller compared to the training set used by \citet{chung-etal-2020-bert}.

% TODO: emphasize, that the meaning is to support teachers, not replace them
\section{Conclusion}
We have collected SweQUAD-MC, the first dataset of Swedish MCQs, and showed the possibility of training usable BERT-based DG models, despite the small scale of the dataset. We have showed that a u-PMLM variant of the BERT-based DG model performs best on the dataset, and proposed a novel methodology of evaluating the plausibility of generated distractors. Around half of the generated distractors were found acceptable by the majority of teachers, and more than 50\% of MCQs had at least one plausible generated distractor, judging by the entropy of students' responses.

Bearing in mind that the aim of the proposed method is to support (not replace) teachers, we deem that our method works well for MCQs in Swedish (and potentially in other languages with a pretrained BERT and a dataset of a similar scale).

Furthermore, we have presented a baseline applicable to any language with a UD treebank (currently about 100 languages). Although its performance is nowhere near the u-PMLM variant, we believe that it can serve as a good point of comparison to emerging neural methods for other languages.

\section*{Acknowledgments}
This work was supported by Vinnova (Sweden's Innovation Agency) within project 2019-02997. We would like to thank the anonymous reviewers for their comments, as well as Gabriel Skantze and Bram Willemsen for their helpful feedback prior to the submission of the paper.
% The acknowledgments should go immediately before the references. Do not number the acknowledgments section.
% \textbf{Do not include this section when submitting your paper for review.}

\bibliographystyle{acl_natbib}
\bibliography{anthology,acl2021}

\appendix
\section{SweQUAD-MC data collection details}
We have used publicly available texts from the websites of Swedish government agencies. The exact list of URLs is provided in the GitHub repository associated with the paper. The exact instructions given to students recruited to collect SweQUAD-MC dataset (and their translation to English) are presented in Figure~\ref{fig:swequad_instructions}. In addition to the given instructions, the students were also given the opportunity to slightly reformulate the distractors found in the text in order to align the syntactic structure with that of the key.

\begin{figure*}[!t]
    \centering
    \begin{tcolorbox}
    
    Imagine that you are a teacher checking reading comprehension skills of your students. Given a text, your task is to create one or more multiple choice questions based on the text, i.e.:
    \begin{enumerate}
        \item formulate a question with the correct answer in the text;
        \item mark the correct answer in the text;
        \item mark some wrong, but plausible options in the text.
    \end{enumerate}
    \vspace{1em}
    When you have written your questions, marked the correct answer (CA) and the wrong alternatives in the text, click on ``Submit''. When you formulate the question, think about the following aspects.
    \begin{itemize}
        \item The question must be independent, i.e., one should not require additional information (on top of the given text) to be able to answer the question.
        \item The question should be unambiguous and have only one possible interpretation.
        \item One should not be able to answer your question without reading the text, which is why even wrong alternatives should be plausible.
        \item Wrong options must be in the same grammatical form as the CA. For instance, if the CA begins with a verb in Past Simple, all wrong options must begin with a verb in Past Simple.
    \end{itemize}
    \vspace{1em}
    Find as many questions as you can (+ the correct answer and wrong alternatives) on each text and then get a new text when you can't find more.
    %It could be possible to find multiple questions on the same text. Write as many questions (+ the correct answer and wrong options) as you can, and click on ``Submit'' after each of them. When you can't find more questions on the text, you can click on the button ``Get new text'' to get a new text.
    \end{tcolorbox}
    \caption{An English translation of the original instructions for SweQUAD-MC data collection (the original instructions in Swedish can be found in the GitHub repository)\vspace{1em}}
    \label{fig:swequad_instructions}
\end{figure*}

\section{Quantitative metrics}
In addition to the metrics 1--10 presented in Section 6.1, we have also looked at the following ones (MCQ\% means ``Percentage of MCQ'' and DIS means ``generated distractor(s)'')
\begin{enumerate}
    \setcounter{enumi}{10}
    \item MCQ\% with at least 1 DIS being capitalized differently from the key
    \item MCQ\% with at least 1 DIS being a distractor from training data.
    \item MCQ\% with at least 1 DIS is in any base text from training data.
    \item MCQ\% with at least 1 DIS appearing in at least 1 base text from training data, but not in their own base text.
    \item MCQ\% with all distractors appearing in the base text.
    \item MCQ\% with all distractors appearing in at least 1 base text from training data.
    \item MCQ\% with all DIS being distractors from training data.
\end{enumerate}

The rationale behind metric 11 was that capitalized answers are named entities and thus one would like distractors also to be named entities. However, it does not always hold. For instance, consider the stem ``Who gets an e-mail with a confirmation of a successful submission of the application for the work permit?'' and the key ``you and your employer''. A distractor ``Migration Agency'' would suit the question perfectly, although capitalization is clearly different.

Metrics 12-17 were candidates to become overfitting indicators. However, metric 2 was excluded, since \emph{AnyDisFromTrainDis} is more informative, given phrases used as distractors in training data can be repeated in other texts. Metrics 13-14 were excluded, since it's unclear whether the higher or lower values are better. For instance, if a text from the training data and the given text are thematically similar, would copying a distractor from training data be considered overfitting? Metrics 15-17 were rejected as too strict, leaving the possibility of actually missing overfitting if only 2 of 3 distractors would meet the criteria.

\section{Model selection}
We have trained both left-to-right and u-PMLM variants for 6 epochs (fixing a random seed for u-PMLM masking procedure to 42). The quantitative performance metrics on the development set for the top-3 models for each variant are presented in Table \ref{tab:ms1}. The best u-PMLM model (i-14000) outperformed the best left-to-right model (i-18000) on most of the quantitative metrics.

\begin{table*}[!htb]
\centering
\begin{tabular}{l|ccc|ccc}
\hline
\multirow{3}{*}{\textbf{Metric}} & \multicolumn{3}{c|}{\textbf{left-to-right}} & \multicolumn{3}{c}{\textbf{u-PMLM}}\\ & \textbf{i-10000} & \textbf{i-14000} & \textbf{i-18000} & \textbf{i-10000} & \textbf{i-14000} & \textbf{i-16000} \\
& \textbf{e-3.02} & \textbf{e-4.23} & \textbf{e-5.43} & \textbf{e-3.59} & \textbf{e-5.02} & \textbf{e-5.74} \\
\hline
M1: DisRecall $\uparrow$ & 9.77\% & 14.29\% & 12.41\% & 17.67\% & \textbf{21.43\%} & 18.80\%\\ 
M2: AnyDisRefMatch $\uparrow$ & 18.25\% & 26.19\% & 21.43\% & 30.95\% & \textbf{37.30\%} & 31.75\%\\
M3: AnyDisInText $\uparrow$ & 64.29\% & 69.84\% & \textbf{73.81\%} & 68.25\% & 72.22\% & 73.81\%\\
\hline
M4: KeyInDis $\downarrow$ & 0.79\% & 1.59\% & 3.17\% & 2.38\% & 5.56\% & 5.56\%\\
M5: AnySameDis $\downarrow$ & 34.13\% & 27.78\% & \textbf{19.84\%} & 9.52\% & 10.32\% & 11.90\% \\
M6: AllSameDis $\downarrow$ & 3.17\% & 1.59\% & \textbf{0.79\%} & 1.59\% & \textbf{0.79\%} & 0.79\%\\
M7: AnyDisRep $\downarrow$ & 0.00\% & 0.00\% & 0.00\% & 0.00\% & 1.59\% & 1.59\%\\
M8: AnyDisEmpty $\downarrow$ & 0.00\% & 0.00\% & 0.00\% & 0.00\% & 0.00\% & 0.00\%\\
\hline
M9: AnyDisFromTrainDis $\downarrow$ & 5.56\% & 5.56\% & 6.35\% & 5.56\% & \textbf{2.38\%} & 2.38\%\\
\hline
M10: MeanNCPTK $\uparrow$ & 0.33 & 0.38 & \textbf{0.39} & 0.41 & 0.41 & 0.41 \\
M11: MedianNCPTK $\uparrow$ & 0.18 & 0.19 & \textbf{0.21} & 0.27 & 0.26 & 0.27 \\
\multirow{2}{*}{M12: ModeNCPTK $\uparrow$} & 1.0 & 1.0 & 1.0 & 1.0 & \textbf{1.0} & 1.0 \\
& (13.3\%) & (18.8\%) & (17.6\%) & (18.1\%) & \textbf{(20.3\%)} & (19.6\%) \\
\hline
\end{tabular}
\caption{\label{tab:ms1} TOP-3 models for left-to-right and u-PMLM variants after model selection on the dev set. i-XXXXX shows a number of iterations since training start, e-X.XX shows a number of epochs corresponding to i-XXXXX. Floating point epochs are due to checkpoints being saved every 2000 iterations.}
\end{table*}

\begin{table*}[!htb]
\centering
\begin{tabular}{r|ccc|ccc|ccc}
\hline
\multirow{2}{*}{\textbf{Metric}} & \multicolumn{3}{c|}{\textbf{i-10000, e-3.59}} & \multicolumn{3}{c|}{\textbf{i-14000, e-5.02}} & \multicolumn{3}{c}{\textbf{i-16000, e-5.74}}\\ 
& \textbf{SF} & \textbf{LF} & \textbf{RND} & \textbf{SF} & \textbf{LF} & \textbf{RND} & \textbf{SF} & \textbf{LF} & \textbf{RND}  \\
\hline
M1 $\uparrow$ & 15.8\% & 13.9\% & 15.8\% & 20.7\% & 14.7\% & 19.9\% & 19.9\% & 15.0\% &	17.7\%\\ 
M2 $\uparrow$ & 25.4\% & 25.4\% & 29.4\% & 36.5\% & 27.8\% & 34.1\% & 34.1\% & 27.0\% & 30.1\%\\
M3 $\uparrow$ & 64.3\% & 63.5\% & 65.9\% & 73.0\% & 66.7\% & 69.8\% & 72.2\% & 66.7\% & 70.6\%\\
\hline
M4 $\downarrow$ & 2.4\% & 2.4\% & 3.2\% & 4.0\%	& 4.8\% & 5.6\% & 4.8\% & 5.6\% & 4.8\% \\
M5 $\downarrow$ & 7.9\% & 11.1\% & 7.9\% & 10.3\% & 9.5\% & 10.3\% & 10.3\% & 8.7\% & 10.3\% \\
M6 $\downarrow$ & 1.6\% & 1.6\% & 1.6\% & 0.8\% & 0.8\% & 0.8\%	& 0.8\% & 0.8\% & 0.8\% \\
M7 $\downarrow$ & 0.0\% & 1.6\% & 0.0\% & 0.0\% & 1.6\% & 1.6\% & 0.8\% & 0.8\% & 3.2\%\\
M8 $\downarrow$ & 0.0\% & 0.0\% & 0.0\% & 0.0\% & 0.0\% & 0.0\% & 0.0\% & 0.0\% & 0.0\%\\
\hline
M9 $\downarrow$ & 5.6\% & 4.8\% & 6.3\% & 4.8\% & 5.6\% & 4.0\% & 4.0\% & 4.0\% & 3.2\% \\
\hline
M10 $\uparrow$ & 0.41 & 0.41 & 0.41 & 0.41 & 0.41 & 0.41 & 0.41 & 0.41 & 0.41 \\
M11 $\uparrow$ & 0.24 & 0.22 & 0.25 & 0.26 & 0.21 &	0.22 & 0.29 & 0.22 & 0.22 \\
\multirow{2}{*}{M12 $\uparrow$} & 1.0 & 1.0 & 1.0 & 1.0 & 1.0 & 1.0 & 1.0 & 1.0 & 1.0 \\
& (18\%) & (17\%) & (19\%) & (20\%) & (18\%) & (20\%) & (19\%) & (18\%) & (19\%) \\
\hline
\end{tabular}
\caption{\label{tab:ms2} Results of model selection by the generation order of distractors for the TOP-3 u-PMLM models.}
\end{table*}

The next experiment concerned the order in which distractors are generated, which we tested only for the best u-PMLM model. We tried generating shortest distractors first (SF), longest first (LF) or in a random order with a fixed seed of 42 (RND). The results of the experiment are presented in Table \ref{tab:ms2}. Evidently, models with SF-generation consistently outperform ones with LF-generation. SF-generation also performs on-par or better than RND-generation. However, fixing a seed is not a generalizable solution, which is why we opted for SF-generation. 
\begin{figure*}[!htb]
	\centering
	\includegraphics[width=0.75\textwidth]{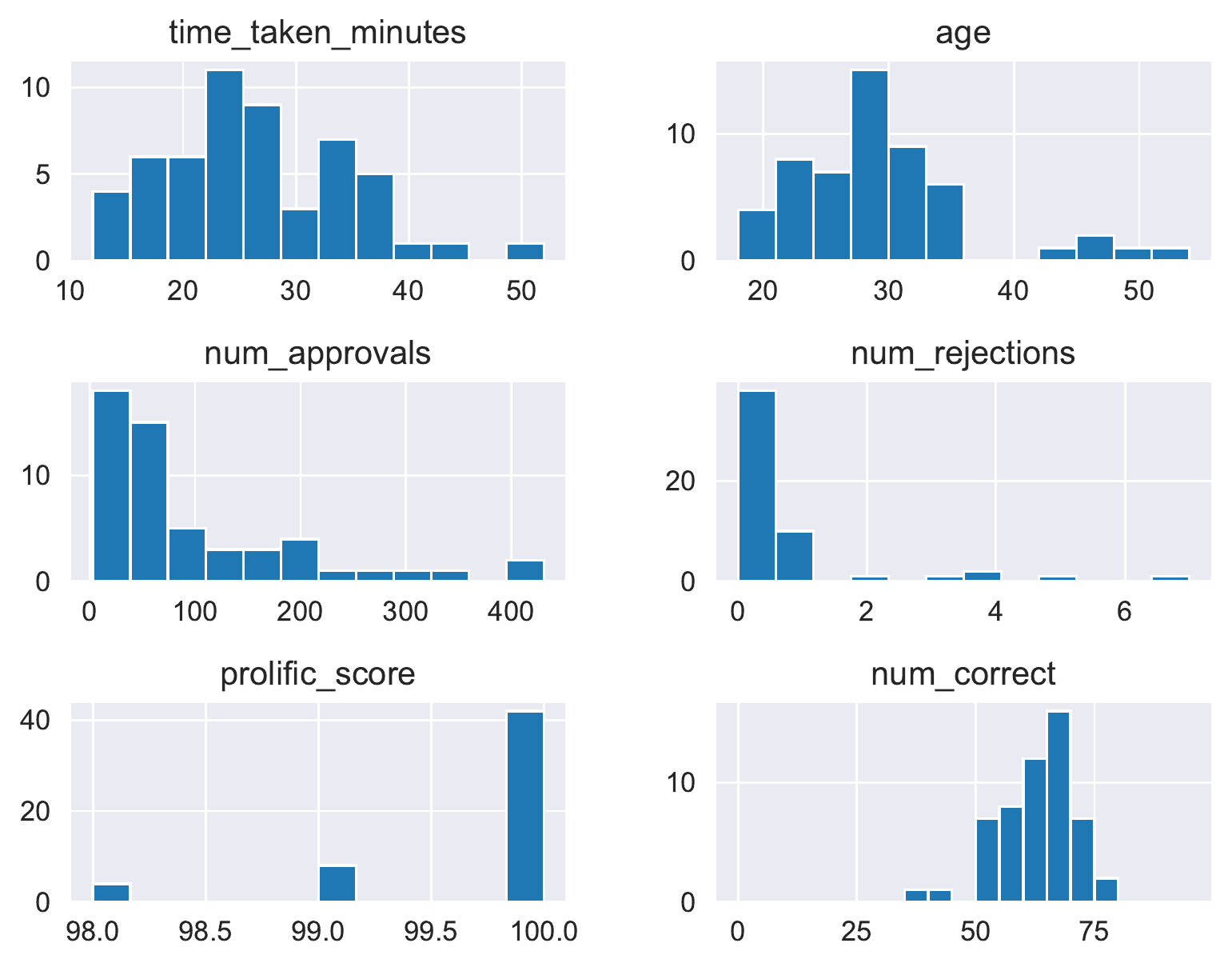}
	\caption{Descriptive statistics of the sample of subjects on Prolific}
	\label{fig:sample_desc}
\end{figure*}
\begin{figure*}[!htb]
    \centering
    % \begin{tcolorbox}
    % Tack för att du deltar i vår undersökning! Du kommer att få se ett antal flervalsfrågor. Din uppgift är att svara rätt på så många av de frågorna som möjligt. Om du inte vet vilket alternativ som är rätt, välj det alternativ som du tycker verkar troligast. Använd \textbf{ENBART} din egen kunskap och sunda förnuft. Använd \textbf{INTE} några andra informationskällor.
    % \end{tcolorbox}
    
    \begin{tcolorbox}
    Thank you for participating in our study! You will be presented with a number of multiple choice questions. Your task is to answer as many of these questions correctly as possible. If you don't know which alternative is correct, choose the one that seems the most plausible. You are allowed to use \textbf{ONLY} your own prior knowledge and common sense. Please, do \textbf{NOT} consult any other external sources of information.
    \end{tcolorbox}
    \caption{An English translation of the original instructions given to subjects on the Prolific platform (the original instructions in Swedish can be found in the GitHub repository)}
    \label{fig:student_eval_instructions}
\end{figure*}

\section{Human evaluation details}
\subsection{Student's perspective}
Evaluation from the student's perspective has been conducted on the Prolific platform\footnote{\href{https://www.prolific.co/}{https://www.prolific.co/}}. We used Prolific's pre-screening feature and required each subject to have Swedish as the first language and hold at least a high school diploma (A-levels). Descriptive statistics about the recruited sample of subjects is presented in Figure \ref{fig:sample_desc}. The exact guidelines given to the subjects (and their translation to English) are presented in Figure \ref{fig:student_eval_instructions}. MCQs were presented in a random order, but the order of options for each MCQs was the same for each subject.

\subsubsection{Check of the t-test assumptions}
\begin{figure}[!b]
	\centering
	\includegraphics[width=0.45\textwidth]{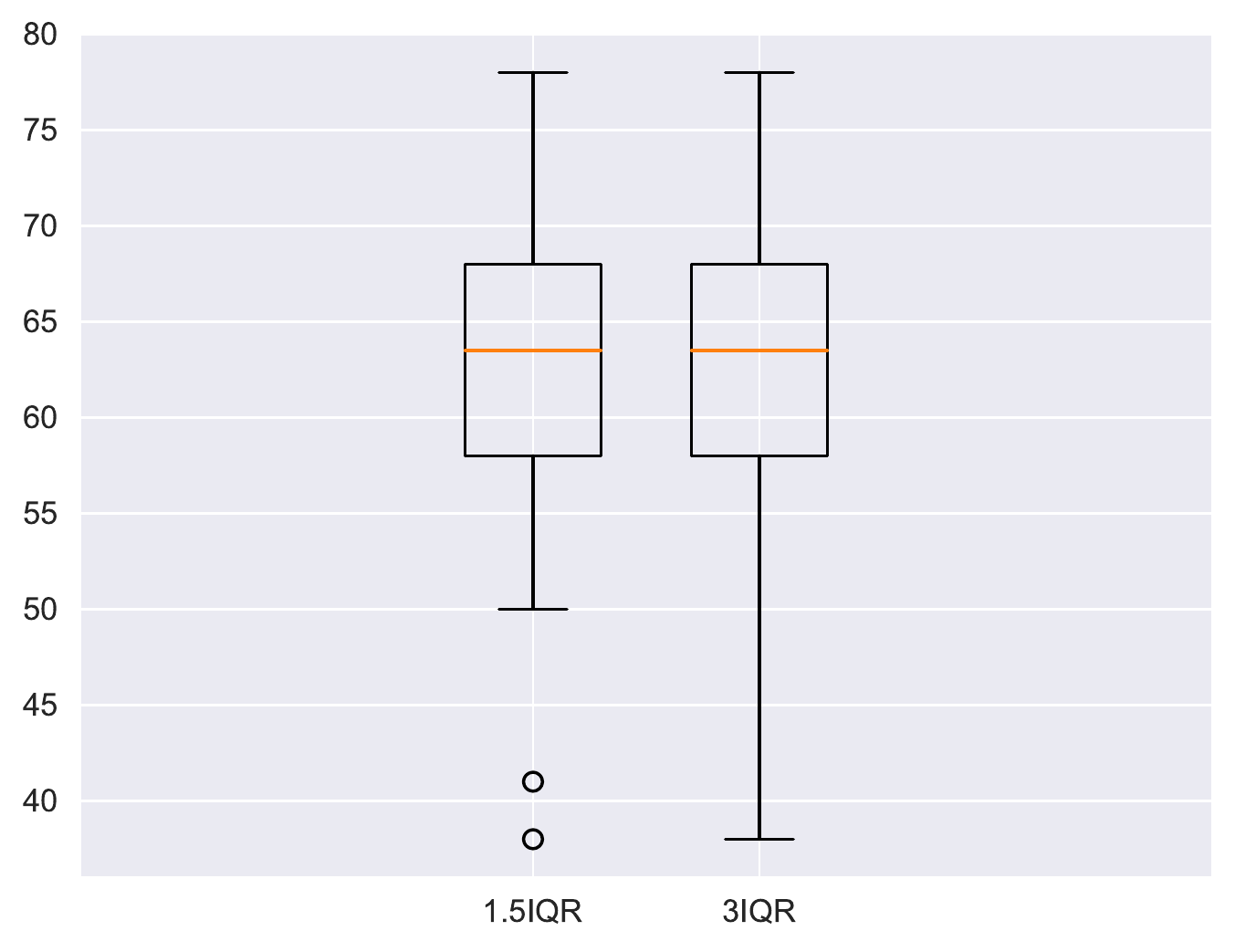}
	\caption{Boxplots for the number of correctly answered questions}
	\label{fig:boxplot}
\end{figure}

We used one sample t-test for conducting our analysis and thus the following assumptions were checked for.
\begin{enumerate}
    \item \textbf{The variable under study should be either an interval or ratio variable}. Our variable, the number of correctly answered MCQs, is clearly on a ratio scale.
    \item \textbf{The observations in the sample should be independent}. Subjects have performed the task independently of each other through a Prolific platform, hence the observations are independent.
    \item \textbf{The variable under study should be approximately normally distributed}. The distribution of the number of correctly answered MCQs is presented in Figure \ref{fig:sample_desc} (the plot in the last row and the last column with the title ``num\_correct''). The distribution is indeed approximately normal.
    \item \textbf{The variable under study should have no extreme outliers}. Outliers are typically defined in terms of the interquartile range (IQR), which equals to Q3 - Q1. The datapoints outside 1.5IQR are deemed mild outliers, whereas those outside 3IQR are considered extreme outliers. Boxplots for our data with whiskers within both 1.5IQR and 3IQR are presented in Figure \ref{fig:boxplot}. Two datapoints can be considered mild outliers, but no extreme outliers are present, which means this assumption for the one sample t-test is not violated.
\end{enumerate}

\subsection{Teacher's perspective}
\subsubsection{Instructions}
The exact guidelines given to the teachers and their translation to English, are presented in Figure \ref{fig:teacher_eval_instructions}.
\begin{figure*}
    \centering
    % \begin{tcolorbox}
    % Tack för att du deltar i vår undersökning! Du kommer att få se ett antal tester. Varje test innehåller en text, en läsförståelsefråga för den texten, det explicit markerade rätta svaret för frågan och ett antal förslag för felaktiga, men troliga alternativ (distraktorer).

    % Antag att du skulle vilja använda  den givna frågan för att testa läsförståelse av den givna texten. Din uppgift är att bedöma vilka av \textbf{de föreslagna distraktorerna} (om någon) som passar bra för just detta. Markera passande distraktorer genom att kryssa i respektive kryssrutor. För de andra distraktorerna (som du inte valde), skriv gärna dina anledningar till varför distraktorerna var olämpliga i respektive textytor (gärna max 1 mening).
    % \end{tcolorbox}
    
    \begin{tcolorbox}
    Thank you for participating in our study! You will be presented with a number of tests. Each test contains a text, a reading comprehension question based on the text, the explicitly marked correct answer to this question and a number of suggestions for wrong, but plausible alternatives (distractors).

    Suppose you would like to use the given question for testing reading comprehension of the given text. Your task is to judge which of the suggested distractors (if any) you would fit the purpose. Select suitable distractors by simply ticking the respective checkboxes. For the other distractors (that you didn't select), please briefly state your reasons why these distractors were inappropriate in the respective text fields (max 1 sentence).
    \end{tcolorbox}
    \caption{An English translation of the original instructions given to teachers (the original instructions in Swedish can be found in the GitHub repository)}
    \label{fig:teacher_eval_instructions}
\end{figure*}

\subsubsection{Inter-annotator agreement}
To evaluate the inter-annotator agreement (IAA) between the teachers, we have reformulated the problem into a ranking problem, where all accepted distractors were given the rank of 1 and those rejected - the rank of 2. IAA was then estimated using Goodman-Kruskal's~$\gamma$ \cite{goodman1979measures}, specifically its multirater version $\gamma_N$ proposed by \citet{kalpakchi2021quinductor}. The total number of concordant and discordant pairs were summed for each pair of teachers for each MCQ. The resulting $\gamma_N$ equals to 0.85, indicating a very large agreement on the scale proposed by \citet{rosenthal1996qualitative}.

\section{Details on surveying related work}
To get a comprehensive overview of methods for generating distractors for MCQs, we employed a two-step process. The first step was to issue queries ``distractor generation'' and ``multiple choice question generation'' to ACL Anthology and Google Scholar. The result was 20 articles from ACL Anthology and 4 additional ones from Google Scholar. The second step was to select relevant references from the ``Related work'' sections of these articles. This resulted into 15 additional articles. Out of found 39 articles, 11 were filtered out (8 focused only on generating questions, 1 relied mostly on expert knowledge, 1 on the auxiliary relation extraction task and 1 was a demo paper), leaving 28 articles in total. Only 2 of these 28 papers worked with a language other than English (Chinese and Basque). 

\section{Generated samples}
A number of generated distractors along with the respective stems and keys from the dataset are presented in Figures \ref{fig:sample1}, \ref{fig:sample2}, \ref{fig:sample3}, \ref{fig:sample4}, \ref{fig:sample5}. The questions are sampled based on the entropy of student's answers using the same 5 buckets as in sampling for teachers' evaluation. Recall that distractors are said to be low frequency (LF-DIS) if they were chosen by less than 5\% of students. Hence, a red cross in the column ``F-DIS $> 5\%$'' entails that a given distractor is in fact an LF-DIS.

The MCQ in sample 1 has an entropy of 0, meaning all students have selected the same option, which in this case was the key. In this case, two of three distractors were accepted by the majority of teachers, although all of them were LF-DIS. This is a good example of an MCQ with plausible distractors, but where the stem is too easy.

The MCQ in sample 2 presents an interesting case, when the distractor contains an obvious grammatical error (comma before the first word in the distractor 3). While the distractor was rightfully rejected by the majority of teachers, it was still selected by more than 5\% of students.

The MCQ in sample 3 is a good example of longer distractors. In this case, two distractors were accepted by teachers and two were selected by more than 5\% of students. However, interestingly these sets are disjoint, meaning that all three distractors could potentially be useful. Another more general observation, requiring future research, is that our model seems to struggle more when generating longer distractors in general, resulting in non-finished sentences or repetitions of words.

The MCQ in sample 4 is somewhat opposite to sample 3, since one distractor that was accepted by the teachers turned out to be an LF-DIS. This either means that the stem was too easy or that none of the distractors were potentially useful.

The MCQ in sample 5 is the one with a highest theoretically possible entropy between selecting the correct or a wrong option. Note that it might still happen that some of the distractors is LF-DIS, since the entropy is calculated not between all four options, but only between the key and the distractors as a group.

% Explain what entropy 0 is
% Explain that those with F-DIS < 5% are LF-DIS
\begin{figure*}
    \centering
    \begin{tcolorbox}
    \textbf{Stem}\\
    Vad täcker över hälften av Sveriges yta?\\
    ({\it What covers more than half of the surface of Sweden?})\\
    
    \textbf{Key:} skog ({\it forest})\\
    
    \begin{tabular}{l|l|P{2cm}|c}
        \textbf{Distractor (sv)} & \textbf{Distractor (en)} & \textbf{Accepted by  teachers?} & \textbf{F-DIS $> 5\%$} \\
        \hline
        vattendrag & {\it water} & \greencheck & \redcross \\
        miljöer & {\it environments} & \redcross & \redcross \\
        djur - och växtarter & {\it plant and animal species} & \greencheck & \redcross \\
    \end{tabular}
    \end{tcolorbox}
    \caption{Sample 1 (entropy 0). ``F-DIS'' denotes the frequency of choice of a distractors by the students, ``Accepted by teachers'' indicates if a distractor was accepted by the majority of teachers.}
    \label{fig:sample1}
\end{figure*}

\begin{figure*}
    \centering
    \begin{tcolorbox}
    \textbf{Stem}\\
    Vad förvaras på en torkanläggning?\\
    ({\it What is stored in a drying facility?})\\
    
    \textbf{Key:} spannmål, hö eller halm ({\it grains, hay or straw})\\
    
    \begin{tabular}{l|l|P{2cm}|c}
        \textbf{Distractor (sv)} & \textbf{Distractor (en)} & \textbf{Accepted by teachers?} & \textbf{F-DIS $> 5\%$} \\
        \hline
        ogräs & {\it weeds} & \greencheck & \redcross \\
        balpressar & {\it balers} & \greencheck & \greencheck \\
        , harvar och sår & {\it , harrows and sows} & \redcross & \redcross \\
    \end{tabular}
    \end{tcolorbox}
    \caption{Sample 2 (entropy 0.31). ``F-DIS'' denotes the frequency of choice of a distractors by the students, ``Accepted by teachers'' indicates if a distractor was accepted by the majority of teachers.}
    \label{fig:sample2}
\end{figure*}

\begin{figure*}
    \centering
    \begin{tcolorbox}
    \textbf{Stem}\\
    När betalar du avgiften om du ansöker på en ambassad?\\
    (When do you pay the fee when you are applying at an embassy?)\\
    
    \textbf{Key:} när du lämnar in din ansökan ({\it when you are handing in your application})\\
    
    \begin{tabular}{p{4.7cm}|p{4.9cm}|P{1.6cm}|c}
        \textbf{Distractor (sv)} & \textbf{Distractor (en)} & \textbf{Accepted by teachers?} & \textbf{F-DIS $> 5\%$} \\
        \hline
        när du får uppehållstillstånd & {\it when you receive your residence permit} & \greencheck & \greencheck \\
        när du ansöker för första gången & {\it when you are applying for the first time} & \redcross & \greencheck \\
        när du ansöker innan tiden för ditt tidigare tillstånd har gått ut & {\it when you are applying before your previous permit has expired} & \greencheck & \redcross \\
    \end{tabular}
    \end{tcolorbox}
    \caption{Sample 3 (entropy 0.57). ``F-DIS'' denotes the frequency of choice of a distractors by the students, ``Accepted by teachers'' indicates if a distractor was accepted by the majority of teachers.}
    \label{fig:sample3}
\end{figure*}

\begin{figure*}
    \centering
    \begin{tcolorbox}
    \textbf{Stem}\\
    Vad är negativt för flera marina miljöer?\\
    ({\it What is negative for several marine environments?})\\
    
    \textbf{Key:} kommersiellt fiske ({\it commercial fishing})\\
    
    \begin{tabular}{l|l|P{2cm}|c}
        \textbf{Distractor (sv)} & \textbf{Distractor (en)} & \textbf{Accepted by teachers?} & \textbf{F-DIS $> 5\%$} \\
        \hline
        klimatförändringar & {\it climate change} &\redcross & \greencheck \\
        många olika miljöer & {\it many different environments} & \redcross & \redcross \\
        äldre brukningsmetoder & {\it older cultivation methods} & \greencheck & \redcross \\
    \end{tabular}
    \end{tcolorbox}
    \caption{Sample 4 (entropy 0.675). ``F-DIS'' denotes the frequency of choice of a distractors by the students, ``Accepted by teachers'' indicates if a distractor was accepted by the majority of teachers.}
    \label{fig:sample4}
\end{figure*}

\begin{figure*}
    \centering
    \begin{tcolorbox}
    \textbf{Stem}\\
    Vilka kan utfärda medicinska rapporter för kabinbesättning?\\
    ({\it Who can issue medical reports for cabin crew?})\\
    
    \textbf{Key:} företagsläkare ({\it company physicians})\\
    
    \begin{tabular}{l|l|P{2cm}|c}
        \textbf{Distractor (sv)} & \textbf{Distractor (en)} & \textbf{Accepted by teachers?} & \textbf{F-DIS $> 5\%$} \\
        \hline
        företagssköterskor & {\it company nurses} & \greencheck & \greencheck \\
        flygläkare & {\it aviation physicians} & \redcross & \greencheck \\
        gymnasieinfo.se & {\it gymnasieinfo.se} & \redcross & \redcross \\
    \end{tabular}
    \end{tcolorbox}
    \caption{Sample 5 (entropy 0.69). ``F-DIS'' denotes the frequency of choice of a distractors by the students, ``Accepted by teachers'' indicates if a distractor was accepted by the majority of teachers.}
    \label{fig:sample5}
\end{figure*}

\end{document}